# The future of AI in critical mineral exploration

Caers[1], J.

[1]Mineral-X, Department of Earth & Planetary Sciences, Stanford University, USA

Corresponding Author: Jef Caers, jcaers@stanford.edu, 450 Jane Stanford Way, Stanford, CA 94305, USA

**Abstract**

The energy transition through increased electrification has put the world's attention on critical mineral exploration. Despite the promise of a growing the demand, the global exploration industry is money-losing enterprise. Even with increased investments a decrease in new discoveries has taken place over the last two decades. In the paper, I propose a solution to this problem where AI is implemented as the enabler of a rigorous scientific method for mineral exploration that aims to reduce cognitive bias & false positives, enhances the role of domain experts and drive down the cost of exploration. The current organization of exploration activities involving many fields of science (geology, geochemistry, geophysics) is no longer an effective in discovering deposits under cover. In particular, the current approach fails to adequately quantify uncertainty, leading to sub-optimal decision-making and $ spent on drilling that are often false positives. Instead, I propose a new scientific method that is based on a philosophical approach founded on the principles of Bayesianism and falsification. In this approach, data acquisition is in the first place seen as a means to falsify human-generated hypothesis. Decision of what data to acquire next is quantified with verifiable metrics and based on rational decision making. A practical protocol is provided that can be used as a template in any exploration campaign. However, in order to make this protocol practical, various form of artificial intelligence are needed. I will argue that the most important form are 1) novel unsupervised learning methods that collaborate with domain experts to better understand data and generate multiple competing geological hypotheses and 2) human-in-the-loop AI algorithms that can optimally plan various geological, geophysical, geochemical and drilling data acquisition where uncertainty reduction of geological hypothesis precedes the uncertainty reduction on grade and tonnage. The approach will be illustrated in several ongoing exploration and resource assessment cases.

# Introduction

The purpose of this paper is to provide a forward-looking essay on the use of AI in critical mineral exploration. This is a personal perspective. The writing is subjective, and contains personal opinions and observations but these are sourced from three decades of unique experience in the use of data science, decision science and AI in the context of Earth resources exploration and development (Caers et al., 2026). In 1998, I wrote a paper (Caers and Journel, 1998) on the use of neural networks trained on outcrop data to enhance geological realism in reservoir models. Aiming to blow everyone out of the water as a very junior Professor, the paper was presented at the Society of Petroleum Engineers Annual Technical Conference and Exhibition in New Orleans while hurricane George was barreling towards Louisiana. Imposing geological realism is essential to accurately predict fluid flow in porous media. Realistic flow modeling ultimately leads to better decision making in the exploration and development of oil/gas fields. In a second paper (Caers and Ma, 2002) on AI for reservoir characterization, in 2002, I demonstrated that neural nets can be used to learn the relationship between borehole and geophysical data, then when applied to areas without boreholes, improves the prediction of important reservoir properties such as porosity. A key contribution was to properly account for the different scale of information between a core in a borehole and a seismic amplitude signal at 1000s ft in the subsurface. This scale issue will be covered at length in this paper.

A decade later a wealth of publication as well as real field implementation emerged on the topic of realistic geological modeling using various types of machine and deep learning (He et al., 2022; J. Nagoor Kani and Elsheikh, 2018; Laloy et al., 2018, 2017; Lewis and Vigh, 2017; Liu et al., 2021; Nasir and Durlofsky, 2023, to list a few). One can hardly imagine oil & gas reservoir engineering today not using various forms of AI, whether it is machine learning, reinforcement learning or AI-based decision making. While oil/gas development based on AI, uncertainty quantification, Bayesian philosophy accelerated in the 21$^{st}$ century, its use in exploration, mining and processing remain sporadic and anecdotal. As a mining engineer graduated in 1996 and re-entering the field in 2019, I have very little changed and that one is still debating the merit of uncertainty quantification in resource modeling. Companies still today use methods such as IDW (inverse distance weighting) developed more than 70 years ago. At the same time the ROI (return on investment) in mining remains abysmal relative to other extraction industries. While oil/gas companies wouldn't dream of building a large offshore platform without quantifying geological uncertainty, and thereby risk, most mines today are still developed using deterministic models. In exploration, deterministic modeling and inversion of multi-physics data is still the norm. Knowledge about the advances that happened regarding quantitative risk assessment in the oil/gas industry appears absent in the minerals industry.

However, things are changing rapidly: over the past decade the global mining industry is now finding itself at the nucleus of the energy transition, with the demand for critical mineral likely increasing, as well as the corresponding geopolitical, social & environmental aspects. This is now generating the formation of new types of companies, often backed by venture capital, who are seeking to disrupt the mining industry, as they already have in healthcare, renewable energy and climate technology. A new wave of digital technology is emerging that may change the current economic, educational, social & industrial systems as we know them through the use of large language models. Whether this actually will have the impact it promises still remains to be seen, but the fear of missing out (FOMO) is ubiquitous. I believe that for the mining industry no fear is warranted, just opportunity in the use of quantitative methods such as artificial intelligence. This paper looks at where those opportunities lie and how AI may contribute.

The work of my group, Stanford Mineral-X has been instrumental to a tier 1 copper discovery in the Zambian Copperbelt (Dempsey, 2024), where AI had a significant impact in how drilling was plan and subsurface models were build. Several start-up companies focusing on AI in mineral exploration have spawned from the program.

This text does not focus on listing possible applications of AI to exploration, nor write an exhaustive review paper as these already exist (Lindi et al., 2024; Yang et al., 2024). Instead, I will view AI from the lens of better decision-making founded on a new scientific method. In part I of this paper, I will lay down the key decision questions that need to be made in exploration and how developing a scientific method to exploration will make the endeavor more efficient. Indeed, today, in mineral exploration, no agreed upon scientific method exists. What exists is a set of tasks executed by domain experts, each with their own scientific method agreed upon as a paradigm within their own fields. I believe that a broader scientific method for the mineral exploration endeavor is largely lacking, and I plan to argue that this is one of the most important issues in making exploration more efficient, meaning less costs to make a discovery. In Part II, I will then review the current most common applications of AI for exploration and point out common errors. Then in Part III I will conjecture what the most likely evolution of AI is in exploration, again through the lens of decision-making.

## Part I: a new scientific method

**What decisions need to be made in exploration?**

The ultimate goal of exploration is to make a new discovery. Discoveries can be made near known deposits (brownfield exploration) or in an unexplored area (greenfield exploration). In the next decade or so, we are likely to see more brownfield discoveries than greenfield ones. Exploration can be seen as an exercise in reducing scale of investigation by reducing uncertainty in knowledge of the existence of economic deposits. Exploration often starts at the continental scale, looking for example at large igneous provinces that may have the favorable tectonic setting where magma (source) can rise to the surface crust (pathway), then get trapped by physical means and chemical precipitation in large and shallow enough concentrations to be economical. At the larger scales, geologist are guided by geological mapping and airborne geophysical data. Anomalies detected on airborne multi-physics surveys are identified (possible by comparison with those of known deposits) to reduce the scale of exploration on the ground by more detailed geological field work, ground geophysical surveys and geochemical assaying of soils. The next steps is by far the costliest: drilling to confirm or falsify hypothesis of a promising geometries/lithologies (e.g. intrusions) and mineralization. In an ideal setting such decisions are made rationally using decision science (Howard, 1968), uncertainty quantification (Scheidt et al., 2018) and understanding the value of information (Eidsvik et al., 2015) to minimize cost by optimally quantifying uncertainty relative to some economic threshold or other property of interest. Various data sources create an understanding of how the mineral system was developed. Data is used to make predictions for regions/zones/location of interest. Data may not just be acquired to directly detect mineralization but to test geological hypothesis. Data may be acquired to reduce uncertainty on orebody geometry. Not only are various decisions made, the objective (reward in AI) of the decision changes during the exploration project.

By decisions we mean the irrevocable allocations of resources ($ amounts), not just mental constructs. Obviously, in exploration, one of the main decisions is: what data should we acquire

next, or do we walk away from the current project. In Part III, I will argue that such rational decision making is largely absent for reasons that have nothing to do with science, instead, with the way the exploration industry is currently organized.

To illustrate the need for a scientific method, I will cover one particular exploration activity: exploration under cover.

**A typical approach to exploration under cover**

The world has seen a decline in new (greenfield) discoveries over the last decade. Not because of lack of investments, in fact they went up in $ values (Schodde, 2017), but because many outcropping deposits have been found. Exploration now needs to go under cover. This means that indirect observations, such as geophysical & geochemical anomalies become increasingly important. A direct observation consist of an observation of mineralization at the surface during geological field work, or during drilling. Here we discuss case where drilling has not been done, and hence, in a promising case, needs to be planned. Here would be a common approach today

- Deterministic inversion of gravity, magnetics and EM into a 3D subsurface models of density, magnetic susceptibility and conductivity.
- Interpretation of the 3D property variation into features/zones of interest (e.g. and intrusion, an alteration, a lithology, a fault).
- Possible follow-up with additional surveys now on the surface of the Earth, then possible re-interpretation of features of interest.
- An expert guided decision on drilling, informed by the deterministically interpreted anomaly, often with the goal to maximize intersection.

While this approach appears logical and certainly has led to successes, my observation in working with companies is that in many cases nothing of interest is intersected when drilling. In other words, a false positive is observed, and worse: the reason why is not understood or evaluated, one simply moves on to the next anomaly. Why is this the case?

The largest problem is determinism. A single geophysical inversion is given a single geological interpretation. Uncertainty at every step of this process is ignored.

In the next section, we will take a very different approach, where the following changes are made

- Deterministic inversion becomes stochastic inversion.
- Interpretation includes a falsification step to eliminate wrong modeling assumptions because they may induce false positives.
- Geological realism is included right from the start, not after inversion. We invert for geology, without vague (non-geological) regularization terms that are typically used in deterministic inversion.
- We propose drilling plans as collaboration between experts and metrics of uncertainty quantification.
- We plan drilling with a given objective in mind (maximal intersection, volume of the intrusion) and do such planning based on quantitative metrics of optimality for the proposed drilling plan.

These changes are not just random choices, they are motivated by a purposeful way of thinking about data, models, objectives and decision that emanate from a long history of scientific approaches. In the next section we will emphasize this through a philosophy of science reasoning, in particular focus on two major philosophies: Falsification & Bayesianism.

**A scientific method for mineral exploration**

*The scientific method*

Humans do science but also study how science is done. This raises philosophical concepts that question how science should be done. What is a good scientific method? The practice of science has evolved over human history starting from empiricism to questions that involve the relationship between models and data. It also considers how humans organize to do science, because various types of social interactions define what standards and norms are, even if they may later be proven incorrect. Insider often don't see faults or epistemic errors, because the community has accepted them as standards. Scientist may be overly optimistic about what they know (Bond et al., 2007). Thomas Kuhn in particular studied these organizations and called them "paradigms". He sees paradigms as consisting of certain (theoretical) assumptions, laws, methodologies, and applications adapted by members of a scientific community (e.g., evolution, plate tectonics, genetics, theory of general relativity). For example, probabilistic methods can be seen as such a paradigm: they rely on axioms of probability and the definition of a conditional probability, maximum entropy, principle of indifference, Monte Carlo simulations, and so on. Researchers within this paradigm do not question the fundamentals of such paradigm. Activities within the paradigm are then puzzle-solving activities (e.g., interpreting faults on a seismic survey) governed by the rules of the paradigm. But often unresolved issues fester under the surface. People who question paradigms are often considered crazy, ignored and outcast from the community. Many such examples in human history. Some end up winning Noble prizes (a recent example is Katalin Karikó, Nobel Prize in Medicine, 2023, who was shunned, ignored, belittled and "not of faculty quality" at the University of Pennsylvania even when her groundbreaking Noble Prize work was already established).

When it comes to mineral exploration there does not seem to be any clearly defined scientific method. What does exist are individual fields of science such as economic geology, geochemistry and geophysics, each with their own organization of puzzle solving activities. While each component is essential, the question begs if the absence of an overarching scientific method to exploration leads to sub-optimal outcomes.

*The Bayesian philosophy*

While Bayes' rule has been around for more than 250 years, the Bayesian approach is only a recent trend in defining a new scientific method. What sets apart Bayesian probability from "regular" probability (in the notion of Pascal) is the use of prior information. Bayes' came up with a problem of gambling where the odds of winning itself was uncertain (certainly true in exploration). There was no true $p$, the probability of winning. Previous scientists, such as Pascal, had tried to solve such problems using classical probability approach such as maximum likelihood estimation of $p$, which relied on defining how likely a dataset would be, if we were to know the truth. However, Pascal's answer to such problem was shown to be biased by Bayes'. Even though we do not know a true $p$, we can conjecture on the plausibility of possible values. This idea is captured quantitatively in the notion of a prior probability model on the parameter $p$ of a gamble. For an unknown chance of success, we may then assume it is between 0 and 1, and we could assume a uniform distribution. Why uniform? That's the subjectivity of the prior but is better than assuming one should a simply estimate a single $p$. Accepting the notion of prior uncertainty is counter-intuitive, yet essential to quantify uncertainty and thereby making optimal decision. Tversky & Kahneman demonstrated through experiments that humans ignore Bayes' and they ended up winning the Nobel Prize for economics in 1973. Yet, my own experience in working with those we do mineral exploration is the avoidance of a prior exactly because it is considered too subjective. They reason that assuming there is only a single true $p$ is considered not subjective! This serious contradiction has hampered the rigorous development of a scientific method for mineral exploration.

Bayes' rule essentially comes down to stating that in order to make predictions with models and data, we need to know two elements: 1) how likely is the new data assuming a given truth 2) how likely is some truth before acquiring the new data. Traditional statistical approaches, such as deterministic inversion only use the first (likelihood).

Let's consider the problem described in Figure 1, where we have a time-domain EM survey and would like to determine electrical properties in the subsurface. The transmitter is a large current loop transmitting EM signal into the Earth, and 19 receivers record vertical component of secondary magnetic field, produced by conductivity anomaly underground, over 12 time channels along a survey line. Only one time channel is shown with a clear anomaly at station 11-14. The surface topography is mostly flat, and the subsurface is discretized into 45*20*45 cells in X, Y, and Z directions, extending 4000m, 2000m, and 4000m in these directions. The depth of discretization is determined by the diffusion distance corresponds to the last time channel.

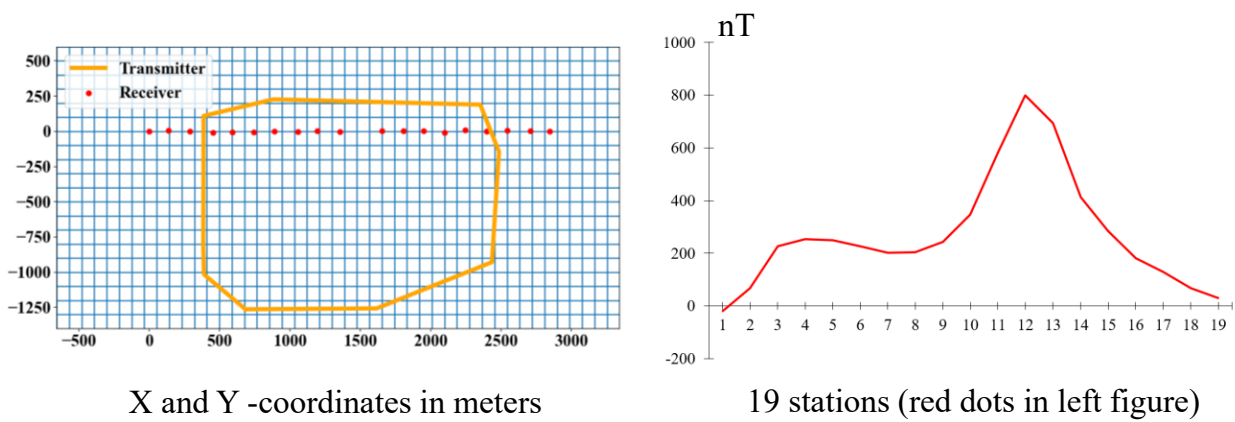

X and Y -coordinates in meters

19 stations (red dots in left figure)

Figure 1: (left) layout of the surface EM survey, with one loop and 19 stations (receivers), (right) one of 16 time channels. The inversion uses all time channels. (courtesy of Zhuo Liu, Stanford University)

The traditional approach is to perform a deterministic inversion, then use that to make a deterministic interpretation, which is shown in Figure 2. Then an expert geologist will interpret the deterministic inversion in terms of shapes, such as intrusions relevant to exploration. Interpretations of the geometry are done in software that focus on a single deterministic model.

With this approach, geophysical inversion happens before geological interpretation. Again, one may argue that the regularization terms help impose geological knowledge, but this is too vague and very limited. In a way the current approach causes a double dip of the same geological knowledge: first we have to impose it in the regularization term, then we impose it once more in the interpretation. It becomes a self-fulfilling prophecy and fraud with circular reasoning. In Bayes' both come together and are solved as a single problem. Bayes' requires one to state plausible solutions (geometries) before doing any inversion.

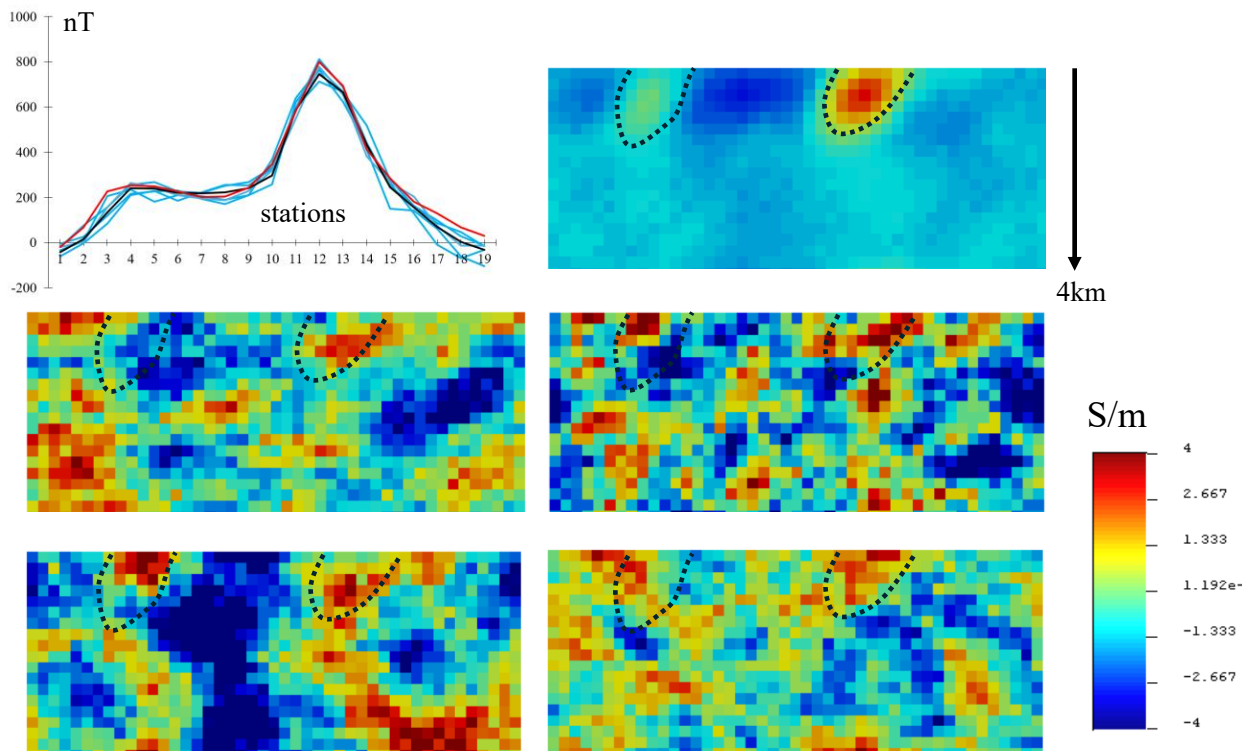

Figure 2: deterministic vs stochastic information. (courtesy of Zhuo Liu, Stanford University). In deterministic inversion, one first inverts the data, then make a geological interpretation. Blue lines are stochastic inversion; red line is deterministic inversion.

Bayesian inversion offers relatively simple solutions to the problem which can be coded up surprisingly easily and are general as well. Let's revisit the case in Figure 2 and discuss how a prior probability model can be designed. We start with the following question: what is the minimum amount of information we have about the subsurface before acquiring any data? First, we have previously observed that petrophysical properties have certain ranges, and if we know what lithologies we are dealing with, we could possibly even get more specific about these ranges. Secondly, we know spatial variation of these properties is not random, nor constant over very large areas. Indeed, we know that the formation of deposits has case very specific variations of these properties. Conceptually, the minimum information is constraints on histograms and the presence of spatial correlation. Mathematically we can translate the latter into a spatial covariance model (in geostatistics: variogram), but we are uncertain about the parameters of this variogram such as the mean, ranges and anisotropy, in fact they could be pretty much anything (except random!). What we have established is a hierarchical way of thinking as follows:

Conceptual idea → high-level mathematical assumptions → detailed mathematical assumptions

which for this case becomes:

The idea of spatial correlation → spatial covariance → parameters of the spatial covariance

Figure 2 now shows that one can generate multiple stochastic inversions, all matching the data equally well as the deterministic inversion, but the spatial variability (variogram) is different for each inversion. Overlaying the deterministically interpreted intrusion, we notice that intrusions may be present in many different shapes, size and orientation, all very different from the single interpretation.

This type of approach can be used in any setting: geologist describe conceptual models (the diagrams found in papers/reports), mathematicians turn these into constraints on the prior probability model, geophysicist use this prior to perform Bayesian inversion. Another example could be:

Norilsk type intrusion → mathematical model of the intrusion shape → parameters of the shape

The most challenging part lies in the first arrow: going from conceptual/descriptive to quantitative. But this is where the largest opportunity lies, and where in Part III we will show AI has much to contribute. But there is a snag in this framework we did not yet address. How do we know that our conceptual idea, and hence mathematical model is not conflicting with the geophysical data or any other data? After all, the initial conceptual idea may be completely incorrect. This type of problem is an example of a wider problem of science: how do we proof that the assumptions we make are correct? It is the Austrian philosopher Karl Popper who spend a lifetime analyzing philosophical questions like these.

*Karl Popper: falsification*

Popper thought that science should not involve induction (theories derived from observations). Instead, theories are seen as speculative or tentative, as created by the human intellect, usually to overcome limitations of previous theories. Once stated, such theories need to be tested rigorously with observations. Theories that are inconsistent with such observation should be rejected (falsified). The theories that survive are the best current theories.

In terms of UQ, one can then see models not as true representations of actual reality but as hypotheses. One has as many hypotheses as models. Such hypothesis can be constrained by previous knowledge, but real field data should not be used to confirm a model (confirmation with data) but to falsify a model hypothesis (reject, the model does not conform with data).

Popper offers a way to address (but not fully solve) the subjectivity of Bayesian reasoning, which is a form of inductive reasoning, as opposed to the deductive form of falsification. If the prior distribution cannot predict the current data (up to data error), then the Bayesian prior is falsified. It basically means that no matter how many models one samples from the prior, each model conflict the data. While Popper's falsification may render the prior less subjective (than before testing), he does not offer a way to revise the prior, that will still be up to human subjectivity.

*Moving forward with a new scientific method*

A scientific method for mineral exploration can be built based on Bayes' and Popper, let's term it a Popper-Bayes approach. In such scientific method, a certain paradigm/protocol can be followed

- Domain scientist generate conceptual model hypotheses, by interpretation of existing data. These are high-level hypotheses, conveyed descriptively

- Experts in numerical modeling turn such high-level hypothesis into a numerical representation that at the highest level has many quantitative hypothesis translated from the conceptual hypothesis but also implement details through model parametrization.
- Model hypothesis are assigned discrete probabilities, while detailed variables require probability distributions.
- This prior distribution model needs to be tested with data. This involves numerical modeling (also termed forward modeling) of the data, done by domain experts of these data sets. A falsification test is a statistical test, which requires statistical metrics and p-values.
- Domain experts in Monte Carlo simulation will provide a Bayesian inversion of the model hypothesis and variables given the data.
- Make decisions based on the Bayesian uncertainty quantification.

To achieve this, all domain experts need to be in the same room. In Part III, we will discuss the various opportunities AI brings to make this paradigm possible, before that, I will review some of the most common application of AI in mineral exploration.

## Part II: current applications of AI

**What is AI?**

To discuss the various forms of AI (Kochenderfer, 2015; Russell and Norvig, 2016; Sutton and Barto, 1998), I will use a specific example instead of formal definitions. Consider a self-driving car. Consider such car given the task of getting from A to B. The goal is to arrive safely & lawfully, within reasonable time, at its destination. To do so, it needs to take various actions of steering, accelerating, decelerating and steering. Sensors classify the current state of the road and predictions are made future actions of itself and other road users or objects. Let's identify the major components

- *Rewards*: the goal: safe, lawful, fast
- *Actions*: steering, braking
- *State*: the current situation on the road with markers, other cars & people, and their direction and velocity

- *Data*: various sensors of the car are used to predict the current state and the next.
- *Belief*: sensors provide imperfect measurement, so the true state is not fully observed, only a belief of the true state.

Solutions to these problems start by formulating the problem into mathematical equations. These equations are essential, and it would be ineffective to right away start solving the problem by collating a set of algorithms. Without going into the details of these equations, we can identify the following mathematical components

- From the current state and actions, a prediction needs to be made on what state we find ourselves in at the next time step. However, actions may not be executed perfectly, so a transition probability model is defined.
- Sensors allows to measure the current state, but sensors are not perfect. This uncertainty (belief in AI) depends on what state we are in. A prediction model, termed belief model, allows updating the current state with new data. Bayes' rule is often used.
- Routes need to be planned and executed, but updates on such plan need to be made. A planning of a route beforehand is called offline planning, while replanning while executing is termed online planning.

Notice that two fundamental uncertainties exists: state transition uncertainty and model (state) uncertainty (belief). Cases with transition uncertainty are often solved using reinforcement learning, while those with model uncertainty are formulated as partially observable Markov decision processes. Partially observed refers to an inexact observation of the true world, while Markov refers to the sequential planning aspect.

Once we have a formulation, we can apply algorithms and tools to solve them

- Predictions can be modeled using machine learning methods, such as neural networks or generative (AI) algorithms.
- Offline and on-line sequential planning under uncertainty can be solved using Monte Carlo Tree search algorithms

Having identified these two major components of AI (predicting and deciding), I will review how they are currently applied in exploration.

*AI that makes predictions*

Making predictions requires datasets. The most important challenge in mineral exploration is the creation and curation of datasets, not the type of machine learning method that is used. Several challenges exist in creating such datasets that can be multi-variate, 3D spatial, compositional and extremal. Rarely does one encounter such challenging prediction/learning problem in other domains. I will consider two crucial and often overlooked or unaddressed components of data curation

- Not all data is on the same scale
- Not all data should be pooled into a single data set
- Learning in area A may not apply to area B

Common errors:

- Upscaling ground truth data (e.g. a core sample) to the geophysical scale.
- Deterministic interpolation of geophysical and geochemical datasets.
- Merging data from different zones cannot be explained geologically.
- Applying machine learning models from area A to Area B without testing the validity of doing so.

All these common errors lead to an increase in false positives, and gives AI a bad name in the industry, while the actual problem is not AI, it is people curating the data. A very significant error in the current paradigm is something that has long been known in mining geostatistics (Journel and Huijbregts, 1978): data at various scales has significant difference in variance. An example is that of comparing interpolated geophysical data with outcrop sampling or core measurement, see Figure 3. Any deterministic interpolation will, with certainty, reduce the variance: the interpolated geophysics has smaller variance than the actual data. Second, label data from drilling at cm scale is often upscaled (or worse, moved) to enable a comparison with the geophysical data. The result is an increased correlation between label and predictor that is induced (not real) because of these operations. As a result, current mineral potential or prospective mapping have a significant number of false positives.

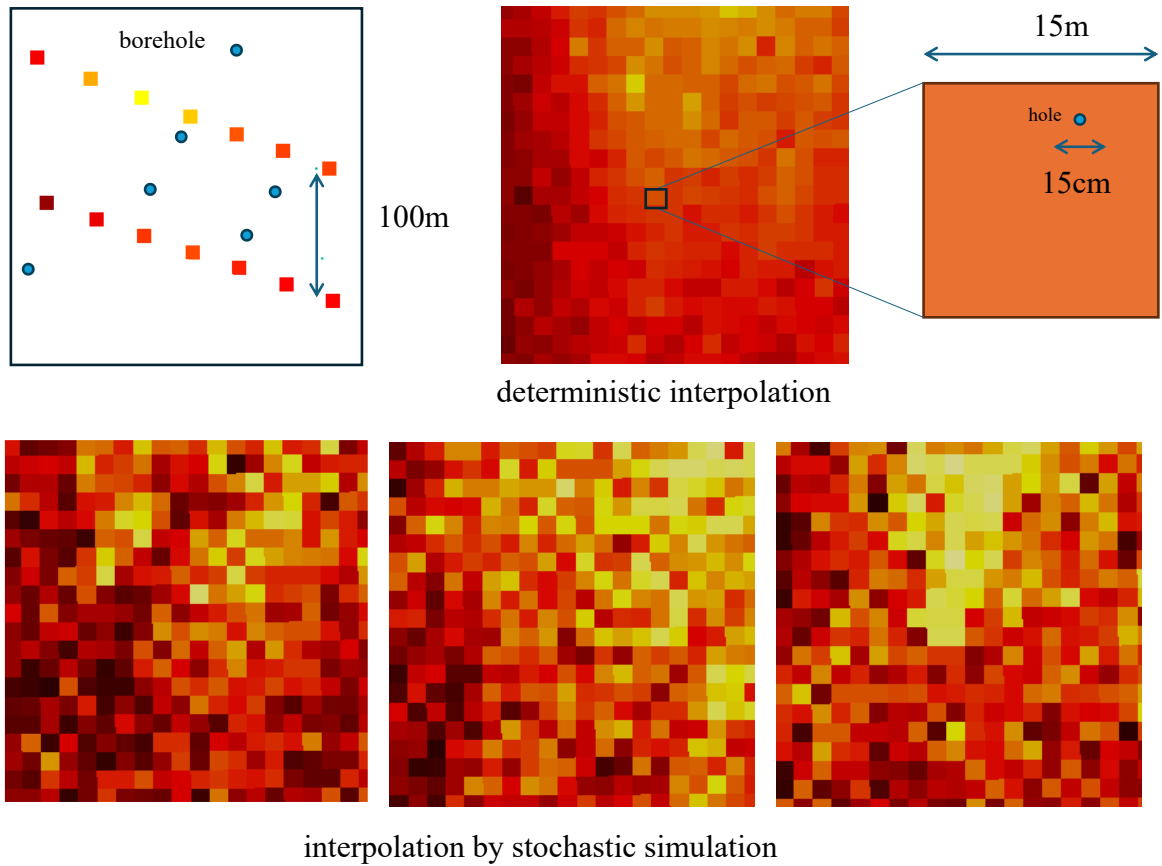

Figure 3: the challenge of having data at different scale and spacing. Geophysical data are collected along lines while borehole data do not necessarily overlap with the geophysical data. Additionally, the scale of information of geophysical and borehole data differ orders of magnitude. Deterministic interpolation has reduced the variance of the data it is interpolating, thereby increasing correlation with borehole information artificially. Stochastic simulation is required to restore this lost variance. Without showing any colorbar, it is clear that the bottom figures are more varied than the top one.

Instead, mineral exploration can be seen as a reduction of uncertainty over multiple scale. Mineral prospectivity maps are informative at the large scale but for the smaller scale additional data through field work is required to narrow down to the prospect scale. Figure shows such as example in Ni-Cu-Co exploration in Canada, where exploration went from the 1000km scale to the 10cm scale. The role of machine learning here can be essential, in the sense that any new information can be used to directly update mineral potential maps, and thereby guide geological field work in real time.

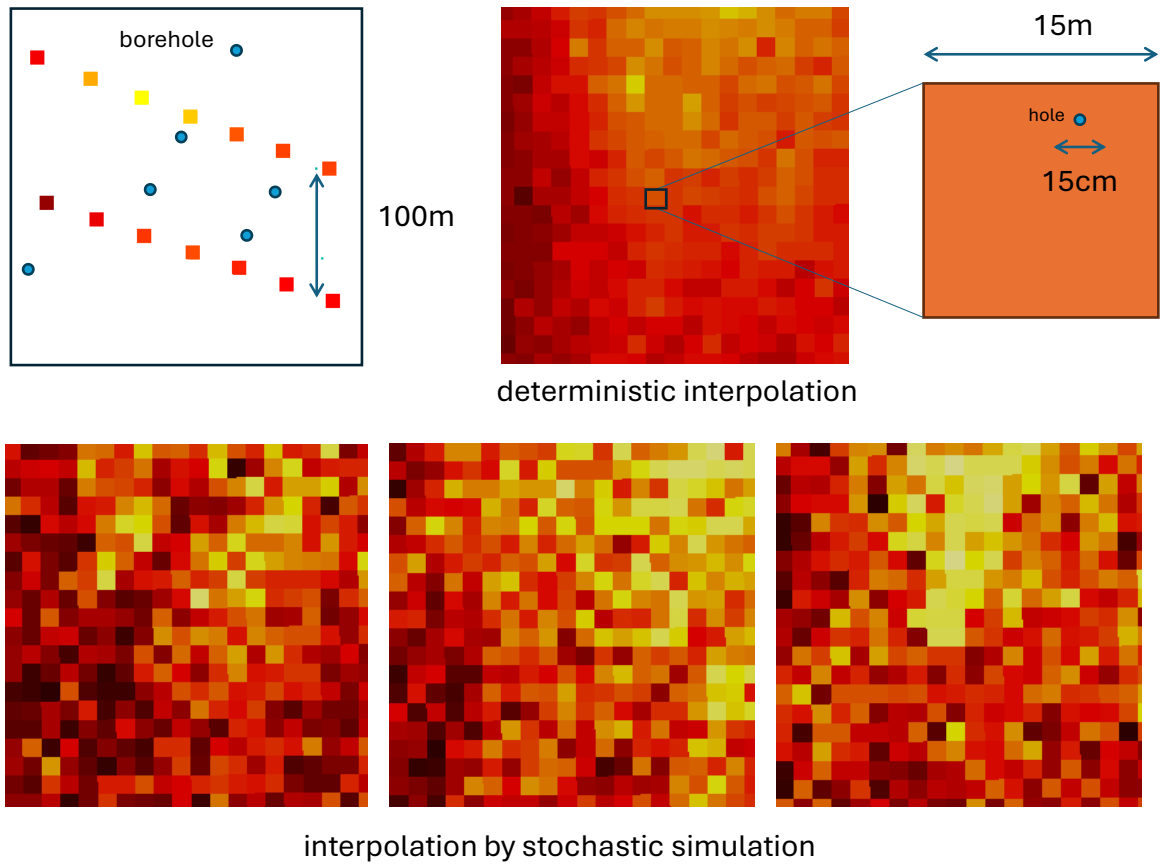

Figure: replica of a real case of predicting concentration of heavy REE measured in boreholes of size 15cm with radiometric data along flightlines that are 100m apart. At the top is shown a deterministic interpolation, whose variance is about a half of the variance of the actual data. Below are three out of many simulated interpolation that respect the data variance and variogram, (including random noise). The interpolation resolution is 15m. All interpolations match the data exactly.

*AI that makes plans*

While machine learning and generative techniques are popular and ubiquitous, the applications AI in decision-making is less common. Indeed, decision-making requires the irrevocable allocation of resources, which is not with the realm of academic research. To date, the only significant and impactful application of decision-making AI in mineral exploration is the characterization of the high-grade portion of the Mingomba deposit Zambia (Dempsey, 2024). In this case historical drilling was re-analyzed using more recent data analysis tools, and new hypotheses were generated that include a higher-grade portion than was previously considered. AI (formulated as a POMDP) was used to sequentially plan drilling. The initial goal of such drilling included the option for the AI to reduce uncertainty on geological hypotheses on the spatial extent of the high-grade mineralization. This method was shown to be much more efficient than the traditional grid-based drilling (Mern and Caers, 2023). The efficiency comes from focusing on falsifying geological hypothesis rather than merely attempt to hit the mineralization.

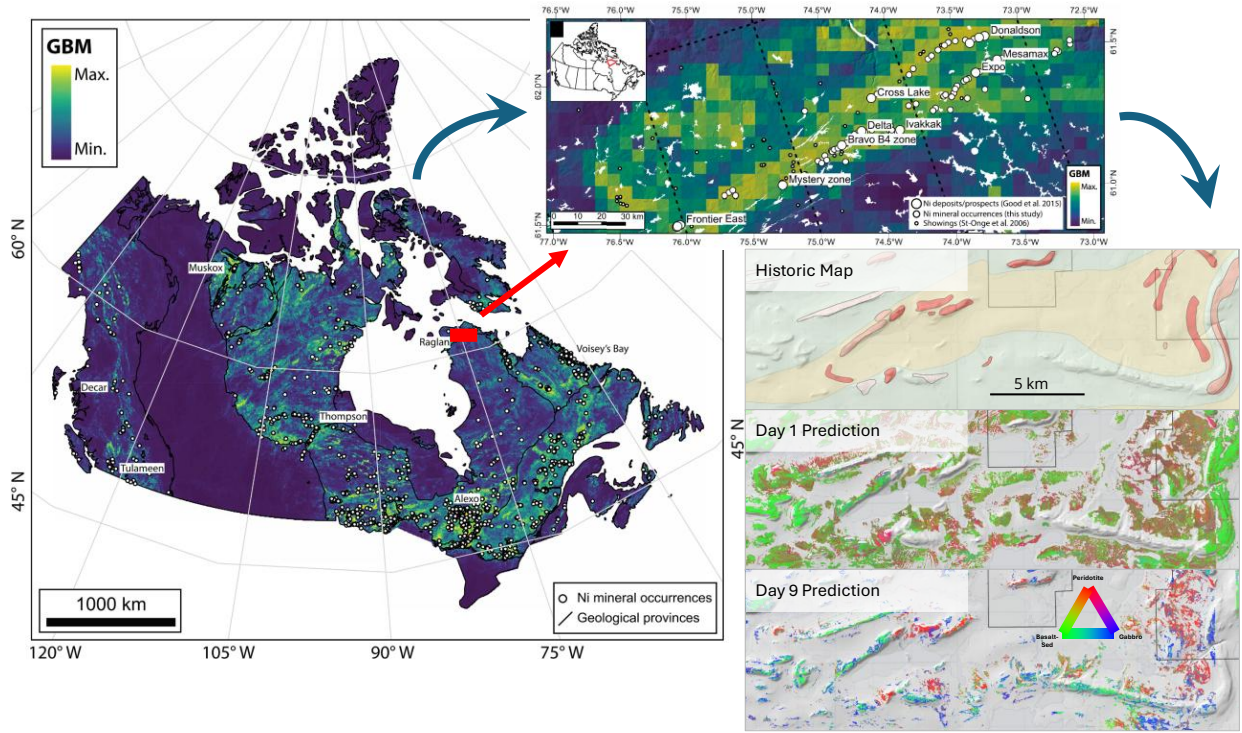

Figure 4: (left) a map showing the potential for Ni-Cu-Co mineralization, the yellow color indicated high potential (from Lawley et al., 2021). (top) a zoom into the Cape smith area, which has high potential. The resolution of the map (the size of the pixel) is 5x5km. (right) AI-assisted geological field work, the resolution of this map is 30cm (multispectral imaging data), allowing geologists to identify efficiently locations of mineralized rocks (red color). Several of the red areas in the day 9 prediction were later confirmed to have Ni-Cu-Co mineralization.

## Part III: the future of AI in mineral exploration

In Part III, I will focus on how AI can help accelerate the discovery and characterization of critical mineral deposits. I will focus first on data analysis and prediction, then on AI-based decision making.

**Human-in-the-loop data science & ML**

The initial deployment of AI was met, in some corners, with a level of defensiveness. Indeed, the perception was that AI would replace or somehow reduce the importance of domain experts. Nobody likes to see their jobs targeted, and this is indeed a common fear today. The perception was that AI would make a new discovery. This will not happen in the way of an AI putting a big red cross on a map where robots will go and drill, make a discovery. The way I view AI in this context is to enhance the domain expert; by removing tasks such expert no longer needs to do or is no good at and focus on the creativity of their process-based thinking (geology and geochemistry).

Let's start with what humans are good at:

- Generate conceptual hypotheses on how mineralization took place from large-scale to deposit scale.
- Interpret relationships in time and space of various geological units.
- Discover patterns in low-dimensional datasets.
- Build complex deterministic numerical models, ensuring geological plausibility.

and what they are not good at

- Read 1000s of documents and create summaries and query systems in a decision-relevant time.
- Discover patterns in high-dimensional multivariate and/or spatial datasets.
- Avoiding cognitive bias.
- Realistic uncertainty quantification.
- Make rational decisions that involve uncertainty.

The most significant potential is an AI-system that interacts with humans and applies novel data science on high-dimensional space-time data to generate plausible hypothesis of mineralization, its extent and grade, and that is decision-ready (e.g. for further data acquisition decisions). And these hypothesis should be falsifiable, meaning that next to stating hypothesis, one should state what data or data outcome will falsify the hypothesis. The latter is critical in avoiding cognitive bias. The decision-ready is essential; deterministic mapping does not provide such readiness.

Foundational to this system are data-science tools. As mentioned above, the most important problem is not the choice of the machine learning algorithm, but the proper curating & thorough analysis of data to generate hypotheses. Humans tend to focus on what they know are indicative relationships of mineralization. For example, in the exploration of massive Ni sulfide deposits geologist use the scatterplot between MgO and Ni to classify the potential for a massive sulfide mineralization. This bivariate plot, possibly supported by a few more, can be enhanced by data science and AI method that may discover additional high-dimensional relationships (four or more) that were previously not discovered. As such, development and use of advanced data science and AI-based or enabled modeling & analysis tools will lead the way in aiding domain experts, for example

- Detection of anomalies on high-dimensional geological & geochemical data.
- High-dimensional pattern analysis of geophysical data to make comparisons with known deposits.
- Non-linear versions of basic data science tools such as PCA, MDS and FA.
- Spatial clustering techniques that can detect complex spatial relationships.
- Advanced geostatistical methods can create complex surfaces constrained to various datasets.
- Data science tool that focus on falsification, instead of data fitting.
- Multi-physics stochastic inversion from borehole to country-scale

This also means that a collaboration between a data scientists and a domain expert is where the magic will happen. We will comment on this topic later.

**AI to support the Popper-Bayes protocol for optimal decision making**

While data science and ML aids in generating hypothesis of locations of interest, decision need to be made on what to do next, mainly, what type of data is now needed, typically

- Dense airborne geophysics
- Surface geophysics & geochemistry
- Drilling

Rational decision-making based on the approach we discussed above means

- Generating multiple high-level geological hypothesis at the deposit scale.
- Generating multiple models of the subsurface constrained to current data, under each geological hypothesis.
- Determining the goal of the drilling, i.e. the reward, or multiple rewards such as better knowing
    - Depth of mineralization
    - Extent of mineralization
    - Maximizing intersection with mineralization

Once multiple hypotheses are generated, metrics of optimality can be used to determine what data should be a acquired next. Efficacy of information (Caers et al., 2022) has been introduced as a metric that quantifies how much future data will reduce uncertainty on average on some quantity of interest. Mathematically this metric compares what we know about the quantity $x$ without the data (the prior probability distribution $f(x)$) and what we would know when we have the data (the posterior distribution $f(m|d)$ for various possible outcomes of the data), the equation goes as follows

When performing sequences of measurements, such as planning a sequence of drilling, one can use AI frameworks such as Monte Carlo tree search to determine the optimal sequential drilling plan (Mern and Caers, 2023). The properties of interest however will change, namely the following hierarchy can be defined. Drill for hypotheses then volume then grade. Indeed, once drilling has falsified all hypothesis but one, focus should be on volume of the mineral system, then the grade, or if desired combine in a grade-tonnage curve.

**Implementation in industry**

Despite the promise of a comprehensive scientific method to make exploration more effective and thereby less costly, several non-scientific barrier exist in the exploration enterprise these are 1) ingrained practices & company organization 3) the exploration industry organization 4) no comprehensive software.

The mining industry remains a conservative industry where innovation is met with resistance. Most large companies focus on making profit from large mining operation and their investment in exploration remains relatively low. Large companies are also hierarchically organized in disciplines, rather than in problem solving. I know of at least one large company that has 1% of experts in data science and 99% in the geosciences.

Companies are organized such that experts in each field handle the responsibilities best suited to their training and experience. The emphasis is often on figuring out details, without knowing whether they in fact affect the ultimate decision. A successful implementation of AI requires a re-organization of large companies around decision-making problems as opposed to task-executing mind-set. This is not simply a matter of re-organization but also a matter of education. Geologists don't need to become data scientists, but there is a great need to learn each other's language, a matter I will revisit later in the education section.

Socialization still happens around a single model, performed with software that is complicit to allow biased & deterministic interpretation. Over time these models are attributed to a sense of truth. As people move on, the reason for the actual interpretation is lost, and the particular geometry created is now considered data, instead of model. Uncertainty in a company is often considered a failure. Not knowing is equated with "lack of knowledge" of a person. If only the person spent more time on the problem and worked harder, uncertainty would be reduced.

Proper software for a rational decision-making approach is lacking. The goal of a software company is to sell software people want to buy it, and those that buy it like software that does what they want it do to, according to the leading paradigm. Rational decision making based on rigorous uncertainty quantification often requires high-performance computing, such as in the implementation of Monte Carlo sampling methods, and come at a $ cost. Small exploration companies are not going to spend tens of thousands of dollars on a cloud computing system, nor will they hire people with the expertise to make this happen.

Finally, most exploration is done using junior mining companies, i.e. small companies owning one or two prospect, employing a few geoscientist. Funding for such operation is often based on showing some form of success (hitting mineralization with one borehole), and the pursuit of additional drilling dollars from investors trumps a rigorous scientific approach to decision making.

Falsifying a geological hypothesis requires long-term thinking in terms of data acquisition, while junior mining companies focus on the short-term only.

It is therefore clear that new financial vehicles are needed. A one-borehole-at-a-time investment is like betting on a single stock each time. However, the common way to invest in stocks is to have a portfolio where risk and return are traded off. The industry should therefore move to investment vehicles that bet on a portfolio of prospects, by understanding the trade-off between risk (uncertainty) and return (estimated present value). How this can be achieved remains to be seen.

**Sustainable decision-making**

Mining has a clear environmental impact and affects the communities that live near these mines. This means that not every economic deposit should be mined if that effect outweighs that impact. Nevertheless, one can easily argue that this impact is inversely proportional to the grade and size of the deposit. Low grade copper mining in the US, now mined at less than 0.5%, require large open-pit mining, while a 6% grade mine in Zambia can be mined using underground mining techniques. Thus, exploration for high-grade deposits will lead to more sustainable mining. Unfortunately, sustainability (social & environmental) is rarely a metric in exploration, and the junior mining company model is not helping either. Sustainability is often a secondary consideration after resources are turned into economic reserves. This may well turn out costly to the company, as construction is delayed to the ensuing lawsuits. Sustainability must therefore be grounded in jurisdiction-specific social and environmental contexts and regulations in federal, state, and local level. Decision-making around exploration should include these additional metrics and preferably should be done of a portfolio of prospects, rather than looking at an individual one. In such portfolio, the risk-return profile of a single deposit now includes all factors and can be traded-off with other prospects.

**Knowledge retention and decimation**

Having been in Academia for more than 30 years, arguably, its largest transformation started in the last few years with the arrival of LLMs. Professor needed to rush to rewrite their syllabus, the way homework and exams were done. I believe this transformation is long overdue. Problems facing humanity such as climate change and the energy transition are far exceeding the breadth of knowledge of a single scientific discipline, yet we still don't offer a degree in Mineral Exploration. Secondly, information is now produced instantaneously, and its intelligence is still improving. Going to the library to copy from a journals is transformed into an algorithm that can search and summarize 1000s of documents, almost in real time.

At the same time, interest in degrees such as economic geology and mining engineering keeps decreasing in the Western world. Now more than ever it is urgent to retain the century-old knowledge about the formation of ore deposits, some stored in dusty geological reports in forgotten closets around the world. Generation of high-quality databases such in Canada and Australia are much needed all over the world. In Africa, only one country has opened such database (gsd.gov.zm). Making such information accessible and open on web platform is essential to attract investment.

## Summary

In this personal essay, I advocate for the scientific and industrial community to come together around a new scientific method for mineral exploration. As exploration is moving "under cover" the existing paradigm of starting with surface outcrops assisted with deterministic geophysical inversion is no longer effective, as indicated by the declining discovery rate. Instead, a decision-focused approach that explicitly assesses risk, through uncertainty quantification will reduce the number of false positive drilling outcomes. This uncertainty quantification will need to acknowledge the fundamental lack of understanding we still have about the nature of orebodies, as modeled through a quantification of epistemic uncertainty. Additionally, proper treatment of data at their respective scales though proper geostatistical method today is lacking and much needed, to reduce false positives.

This change will not happen overnight, as the industry and parts of academia remain set on socializing decision on a single conceptual and 3D model. It will require reorganization of the industrial apparatus, new funding models and new form of education that focus on problem solving in addition to the useful domain expertise focus.

## Acknowledgments

I thank members of the Mineral-X industrial affiliates program for funding this work.

## Statement and Declarations

Conflict of Interests: the authors declare that they have no known competing financial interests or personal relationships that could have appeared to influence the work reported in this paper.

## References

Bond, C.E., Gibbs, A.D., Shipton, Z.K., Jones, S., 2007. What do you think this is? "Conceptual uncertainty" in geoscience interpretation. GSA Today 17, 4. https://doi.org/10.1130/GSAT01711A.1

Caers, J., Journel, A.G., 1998. Stochastic Reservoir Simulation Using Neural Networks Trained on Outcrop Data. Presented at the SPE Annual Technical Conference and Exhibition, New Orleans, Louisiana.

Caers, J., Ma, X., 2002. Modeling Conditional Distributions of Facies from Seismic Using Neural Nets. Mathematical Geology 34, 143–167. https://doi.org/10.1023/a:1014460101588

Caers, J., Scheidt, C., Li, L., 2026. Decision-making uncer uncertainty in developing Earth resources, AGU-Wiley. ed.

Caers, J., Scheidt, C., Yin, Z., Wang, L., Mukerji, T., House, K., 2022. Efficacy of Information in Mineral Exploration Drilling. Nat Resour Res 31, 1157–1173. https://doi.org/10.1007/s11053-022-10030-1

Dempsey, H., 2024. Bill Gates-backed mining company discovers vast Zambian copper deposit. Financial Times.

Eidsvik, J., Mukerji, T., Bhattacharjya, D., 2015. Value of Information in the Earth Sciences: Integrating Spatial Modeling and Decision Analysis. Cambridge University Press.

He, J., Tang, M., Hu, C., Tanaka, S., Wang, K., Wen, X.-H., Nasir, Y., 2022. Deep Reinforcement Learning for Generalizable Field Development Optimization. SPE Journal 27, 226–245. https://doi.org/10.2118/203951-PA

Howard, R.A., 1968. The Foundations of Decision Analysis. IEEE Transactions on Systems Science and Cybernetics 4, 211–219. https://doi.org/10.1109/tssc.1968.300115

J. Nagoor Kani, Elsheikh, A.H., 2018. Reduced-Order Modeling of Subsurface Multi-phase Flow Models Using Deep Residual Recurrent Neural Networks. Transport in Porous Media 126, 713–741. https://doi.org/10.1007/s11242-018-1170-7

Journel, A.G., Huijbregts, C.J., 1978. Mining Geostatistics. The Blackburn press.

Kochenderfer, M.J., 2015. Decision Making Under Uncertainty: Theory and Application. MIT Press.

Laloy, E., Hérault, R., Jacques, D., Linde, N., 2018. Training-Image Based Geostatistical Inversion Using a Spatial Generative Adversarial Neural Network. Water Resources Research 54, 381–406. https://doi.org/10.1002/2017wr022148

Laloy, E., Hérault, R., Lee, J., Jacques, D., Linde, N., 2017. Inversion using a new low-dimensional representation of complex binary geological media based on a deep neural network. Advances in Water Resources 110, 387–405. https://doi.org/10.1016/j.advwatres.2017.09.029

Lawley, C.J.M., Tschirhart, V., Smith, J.W., Pehrsson, S.J., Schetselaar, E.M., Schaeffer, A.J., Houlé, M.G., Eglington, B.M., 2021. Prospectivity modelling of Canadian magmatic Ni (±Cu ± Co ± PGE) sulphide mineral systems. Ore Geology Reviews 132, 103985. https://doi.org/10.1016/j.oregeorev.2021.103985

Lewis, W., Vigh, D., 2017. Deep learning prior models from seismic images for full-waveform inversion. Presented at the SEG Technical Program Expanded Abstracts 2017, Society of Exploration Geophysicists. https://doi.org/10.1190/segam2017-17627643.1

Lindi, O.T., Aladejare, A.E., Ozoji, T.M., Ranta, J.-P., 2024. Uncertainty Quantification in Mineral Resource Estimation. Natural Resources Research 33, 2503–2526. https://doi.org/10.1007/s11053-024-10394-6

Liu, B., Yang, S., Ren, Y., Xu, X., Jiang, P., Chen, Y., 2021. Deep-learning seismic full-waveform inversion for realistic structural models. GEOPHYSICS 86, R31–R44. https://doi.org/10.1190/geo2019-0435.1

Mern, J., Caers, J., 2023. The Intelligent Prospector v1.0: geoscientific model development and prediction by sequential data acquisition planning with application to mineral exploration. Geoscientific Model Development 16, 289–313. https://doi.org/10.5194/gmd-16-289-2023

Nasir, Y., Durlofsky, L.J., 2023. Deep reinforcement learning for optimal well control in subsurface systems with uncertain geology. Journal of Computational Physics 477, 111945. https://doi.org/10.1016/j.jcp.2023.111945

Russell, S.J., Norvig, P., 2016. Artificial intelligence : a modern approach. Pearson.

Scheidt, C., Li, L., Caers, J., 2018. Quantifying Uncertainty in Subsurface Systems, Geophysical Monograph Series. John Wiley & Sons.

Schodde, R., 2017. Recent trends and outlook for global exploration. Presented at the International Convention, Trade Show & Investors Exchange, Toronto, Canada, pp. 5–8.

Sutton, R.S., Barto, A.G., 1998. Reinforcement Learning: An Introduction. MIT Press.

Yang, F., Zuo, R., Kreuzer, O.P., 2024. Artificial intelligence for mineral exploration: A review and perspectives on future directions from data science. Earth-Science Reviews 258, 104941. https://doi.org/10.1016/j.earscirev.2024.104941

2626